\documentclass[runningheads]{llncs}
\usepackage[T1]{fontenc}
\usepackage{amsmath,amssymb,amsfonts,booktabs}
\usepackage{graphicx}
\usepackage{multirow}
\usepackage{xcolor}
\usepackage[font=small,skip=4pt]{caption}
\begin{document}
\title{Beyond Reconstruction: A Physics Based Neural Deferred Shader for Photo-realistic Rendering\protect\footnote{This paper is accepted for the 34th International Conference on Artificial Neural Networks (ICANN), in Kaunas (Lithuania), September 2025.}}
\titlerunning{Physics Based Neural Deferred Shader}
\author{Zhuo He\inst{1} \and
Paul Henderson\inst{1}\orcidID{0000-0002-5198-7445} \and Nicolas Pugeault\inst{1}\orcidID{0000-0002-3455-6280}}
% \author{Anonymous submission}
%
% \authorrunning{Zhuo He et al.}
\institute{School of Computing Science, University of Glasgow, \\
Glasgow G12 8RZ, United Kingdom}

% \institute{Anonymous submission}

\renewcommand{\paragraph}[1]{\par\noindent\textbf{#1}~}

\maketitle              % typeset the header of the contribution
\begin{abstract}
Deep learning based rendering has achieved major improvements in photo-realistic image synthesis, with potential applications including visual effects in movies and photo-realistic scene building in video games. However, a significant limitation is the difficulty of decomposing the illumination and material parameters, which limits such methods to reconstructing an input scene, without any possibility to control these parameters. This paper introduces a novel physics based neural deferred shading pipeline to decompose the data-driven rendering process, learn a generalizable shading function to produce photo-realistic results for shading and relighting tasks; we also propose a shadow estimator to efficiently mimic shadowing effects. Our model achieves improved performance compared to classical models and a state-of-art neural shading model, and enables generalizable photo-realistic shading from arbitrary illumination input.

\end{abstract}
\section{Introduction}
\label{sec:intro}
Photo-realism is a key goal of computer graphics~\cite{newell_progression_1977}, and is essential for creating immersive multimedia experiences and imagery indistinguishable from the real world. Significant advancements have been achieved in illumination representation, material appearance description, and high-precision geometric modeling~\cite{deering_triangle_1988,dick2009efficient,guarini2024pbr,walter_microfacet_2007}. These techniques can be defined as \textit{forward approaches}, that study and model physical interactions between light and surfaces in the real world. However, they require approximations of complex real-world physics and calculations that can reduce the rendering fidelity; there still remain perceptible differences between synthesized and real-world images.

In contrast, neural network-based approaches allow learning aspects of the rendering process from real-world data; these can be seen as \textit{inverse approaches} or \textit{data-driven approaches}, which reconstruct 3D information from 2D images. For example, Neural Radiance Fields \cite{mildenhall_nerf_2020} regress a scene radiance distribution with a neural network by estimating color and density of sample points, then using volume rendering to construct images from novel viewpoints. Gaussian Splatting \cite{kerbl_3d_2023} uses lightweight 3D Gaussians as the primitive representation, with a fast approximate rasterization method to render these to pixels efficiently. 
Although these approaches can produce highly realistic images, they fit models that are specific to the content of individual scenes. 
They must be retrained from scratch for new scenes, and do not learn information that transfers or generalizes across scenes.
\begin{figure}
    \centering
    \includegraphics[width=\textwidth]{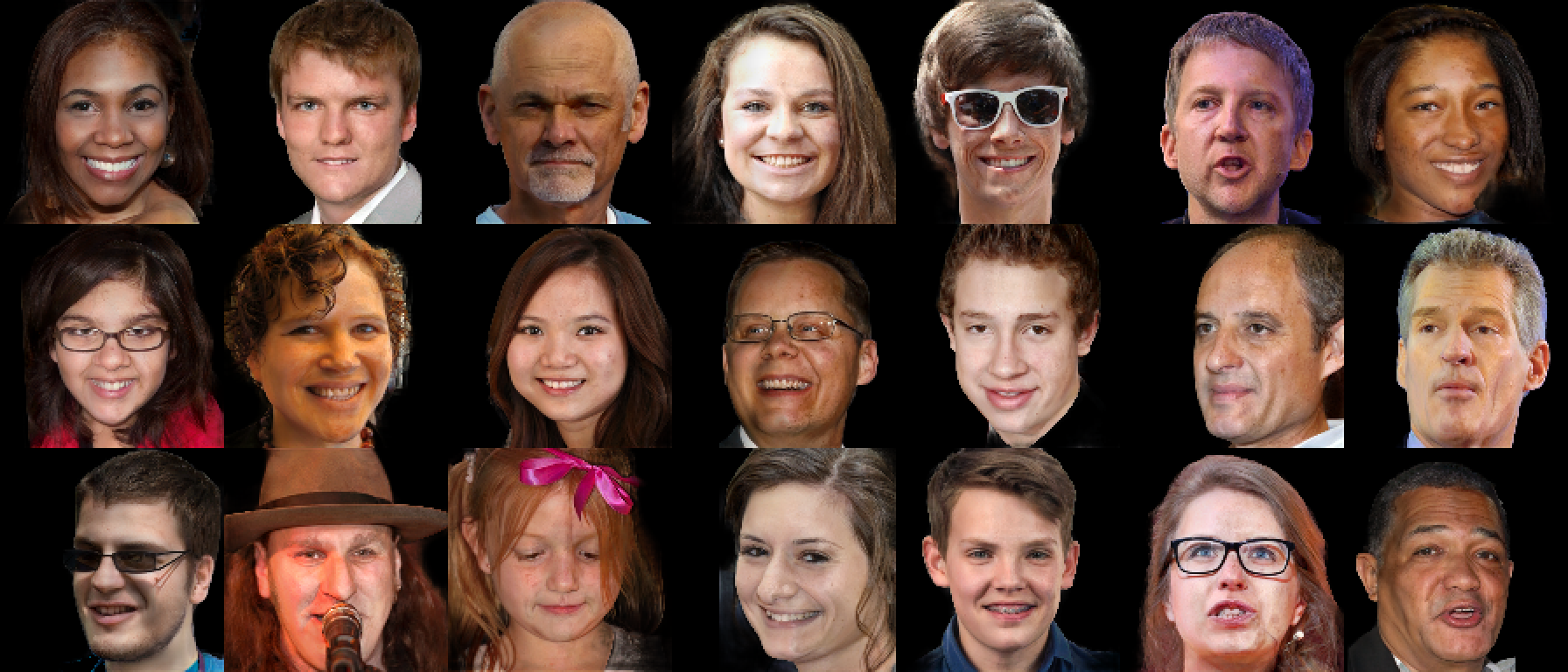}
    \caption{Example of rendering results from our physics based neural deferred shader.}
    \label{fig:fig01}
\end{figure}

Another approach to photorealistic rendering is to use an auxiliary enhancement network to learn a mapping between the domain of traditionally rendered images and that of photorealistic images \cite{kim_switchlight_2024,richter_enhancing_2023}. As the initial rendering result lacks complete 3D scene information such as spatial distribution of illumination and 3D object positions, it is difficult for such post-hoc methods to learn a mapping to photorealistic images. In this work, we combine a deferred shading framework with a neural shading network, trained to directly regress pixel color from physically-based rendering (PBR) textures and light input, to achieve high fidelity rendering results. Our contributions are as follows:
\begin{enumerate}
  \item We propose a physics-based neural deferred shading pipeline that renders scenes with PBR textures (albedo, roughness, specular) and illumination (HDRI light map) to photo-realistic images (Fig.~\ref{fig:fig01}).
  \item We develop a neural shadow estimator to efficiently approximate realistic shadows, improving the final rendering result.
  \item We propose a new FFHQ256-PBR dataset which contains RGB images, material textures (albedo, roughness, specular), depth, screen space ambient occlusion and HDRI environment map.
  % \item We use state-of-the-art texture and illumination estimation methods to obtain a ground-truth dataset of rendering inputs for real-world images, and propose a new FFHQ256-PBR dataset with RGB images, material textures (diffuse, roughness, specular), depth maps, and HDRI environment maps.
  \item We conduct comprehensive experiments to compare our approach with existing shading models on multiple datasets, demonstrating its superior performance.
\end{enumerate}

\section{Related Work}

%Recent advance achieves many successes on generating photo-realistic images of 3D environments using neural networks, it becomes essential for the development of immersive 3D applications. Present methodologies typically either reconstruct a photo-realistic 3D scene that can simulate the novel view rendering or to enhance traditional rendering techniques, others incorporate traditional rendering techniques into learning-based framework. We discuss each of these approaches below.

\paragraph{Photo-realistic Novel View Synthesis.}
Novel view synthesis is an application of the 3D reconstruction task, and recent learning-based techniques allow to reconstruct photo-realistic 3D scenes from posed images, and to synthesize high quality novel views from learned 3D information. Specifically, a series of works on Neural Radiance Field~\cite{guo_nerfren_2022,mildenhall_nerf_2020,szymanowicz_splatter_2024,yu_pixelnerf_2021} represent a single 3D scene using an implicit radiance field function, it samples points along camera rays, regressing color and density to reconstruct the view image. Other works on Gaussian splatting~\cite{kerbl_3d_2023,szymanowicz_splatter_2024,ye_3d_2024,yu_mip-splatting_2024}, on the other hand, draw inspiration from point-based rendering to use a splatting procedure to render pixel colors. They use lightweight 3D Gaussians as the primitive representation and achieve real-time novel view synthesis without sacrificing quality. Both of those approaches require overfitting to specific scenes and are difficult to generalize since they compose geometry, material and illumination together. In contrast, our work is based on the standard deferred rendering pipeline which learns the shading process that can render the arbitrary scene and producing photorealistic results.

\paragraph{Photorealism Enhancement.}
Other works reduce the photo-realism gap between the traditional rendering output and real-world imagery by designing an enhancement model. Richter et al.~\cite{richter_enhancing_2023} capture all shader inputs of the video game Grand Theft Auto 5 (GTA5) and design an adversarial learning process to improve the vividness of shading result by discriminating it with the real-world photo in a similar scene, enhancing the photorealism of render outcome. 
Kim et al.~\cite{kim_switchlight_2024} use a neural network to improve the effect of the rendering outcome from a physically-based Cook-Torrance model, it does not use adversarial learning as it has the paired input and targets. However training a model to improve forward approaches (see Sec.~\ref{sec:intro}) requires either unstable adversarial learning or is restricted by the limited input, as the initial rendering result has discarded much information, like the illumination angle, which would be valuable for the enhancement process. 
In contrast, we propose a more general process of shading, where a neural network regressor provides an atomic model of the shading process, calculating the RGB color of each rasterized shading point from highly informative PBR texture and illumination input, achieving better performance and explainability.

\paragraph{Neural Deferred Rendering.}
\label{sec:NDR}
In classical rendering, there are two ways to render a spatial point: forward shading and deferred shading. Forward shading directly renders each object in turn, and lights it according to all light sources in the scene. Deferred shading instead delays intensive rendering tasks, such as shading, to a later stage, only processing them for the visible points on camera screen, reducing the calculation complexity~\cite{deering_triangle_1988}. Neural rendering is typically based forward shading, albeit learning to represent a scene from sample imagery~\cite{boss2021neuralpil,hadadan_neural_2021,tewari_advances_2022,yao_neilf_2022,zeltner_real-time_2024}, however the deferred pipeline can also be used in neural rendering, which is advantageous as it avoids the non-differentiable rasterization process in classical rendering, and allows replacing classical shading formulae with a neural model learnt from data. Neural deferred rendering was first proposed in Thies et al.~\cite{thies_deferred_2019}, which integrates the conventional real-time rendering pipeline with a learnable neural texture and a deferred neural renderer. Worchel et al.~\cite{worchel_multi-view_2022} extend the deferred neural renderer to tackle 3D reconstruction, using a differentiable rasterizer in the classical deferred rendering pipeline to reconstruct a triangle mesh from multi-view input. However, existing works on neural deferred shading do not consider the light-surface interaction during shading, and are not physics-based---thus unable to generalize to new illumination conditions. Our work redesigns the coordinate-based neural deferred shader~\cite{worchel_multi-view_2022} by incorporating the inbound light and bidirectional angle, regressing the shading result based on physically-meaningful information.

\section{Method}
Image rendering aims to model how light interacts with surfaces \cite{kim_switchlight_2024}, as described by the rendering equation:
\begin{equation}
    L_o(\mathbf{v}) = \int_{\Omega} F(\mathbf{v},\mathbf{l})L_i(\mathbf{l}) \langle \mathbf{n} \cdot \mathbf{l} \rangle d\mathbf{l}
    \label{eqn:eqn01}
\end{equation}
where $L_o(\mathbf{v})$ is the outbound radiance leaving in direction $\mathbf{v}$; it is the integral of the incident light $L_i(\mathbf{l})$ from every possible direction $\mathbf{l}$ across the hemisphere $\Omega$, centered around the surface normal $\mathbf{n}$. $F(\mathbf{v}, \mathbf{l})$ is the Bidirectional Reflectance Distribution Function (BRDF) describing how the surface reflects light.
\begin{figure*}[t]
    \centering
    \includegraphics[width=1.0\textwidth]{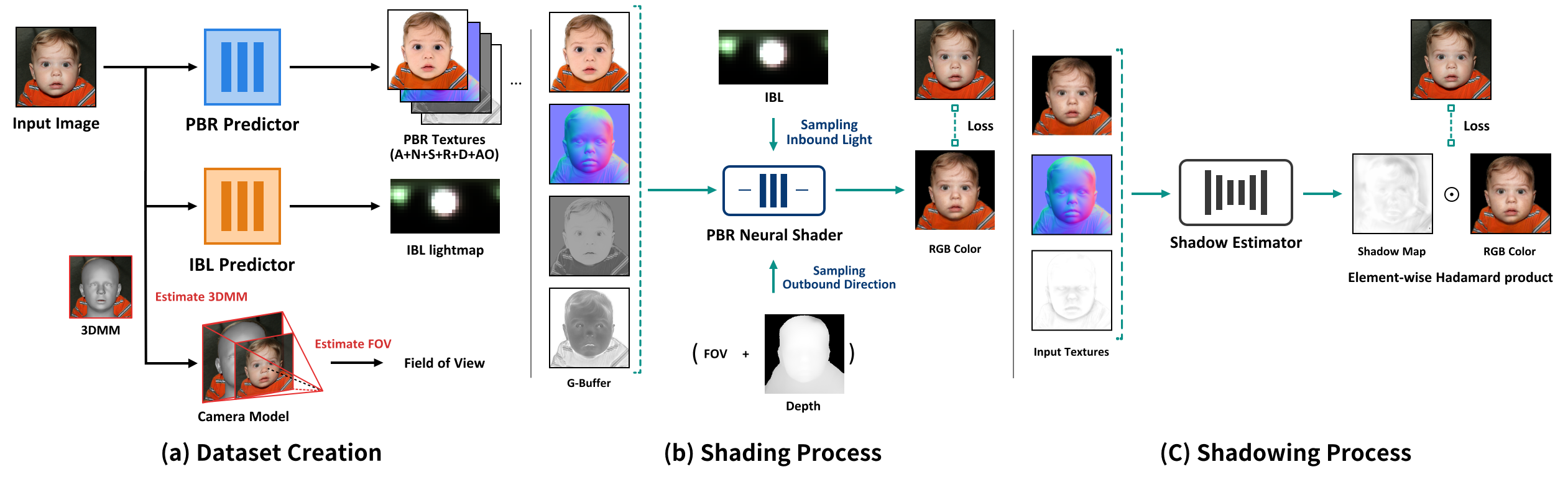}
    \caption{{The overall pipeline of our physics-based neural deferred shading. In data preprocessing (a), the input image is estimated for PBR textures (A: albedo, N: Normal, S: specular, R: roughness, D: depth, AO: Ambient Occlusion), IBL lightmap, field of view via pre-trained models. Then the estimated data are used to train the physics based neural shader (b). Subsequently a shadowing estimator (c) estimates the shadow map applied to the final shading result.}}
    \label{fig:fig02}
\end{figure*}

In practice, approximations of this equation are used, limiting the photorealism of the rendering.
%As the extreme complexity of light-surface interaction in real world, there are many approximations to design each term of the rendering equation in classical way, limiting the photorealism of the result.
We introduce \textit{physics-based neural deferred shading} in this work, a novel framework for photorealistic rendering applied to human portraits. We use a neural network to regress the rendering equation with geometry, material and light inputs instead of using hand-coded approximations. An overview of our pipeline is presented in Fig.~\ref{fig:fig02}. We introduce the main approach in Sec.~\ref{sec:pbnds_method}, then describe the loss functions in Sec.~\ref{sec:loss}, and provide implementation details in Sec.~\ref{sec:implementation}.
\begin{figure}[t]
    \centering
    \includegraphics[width=0.75\textwidth]{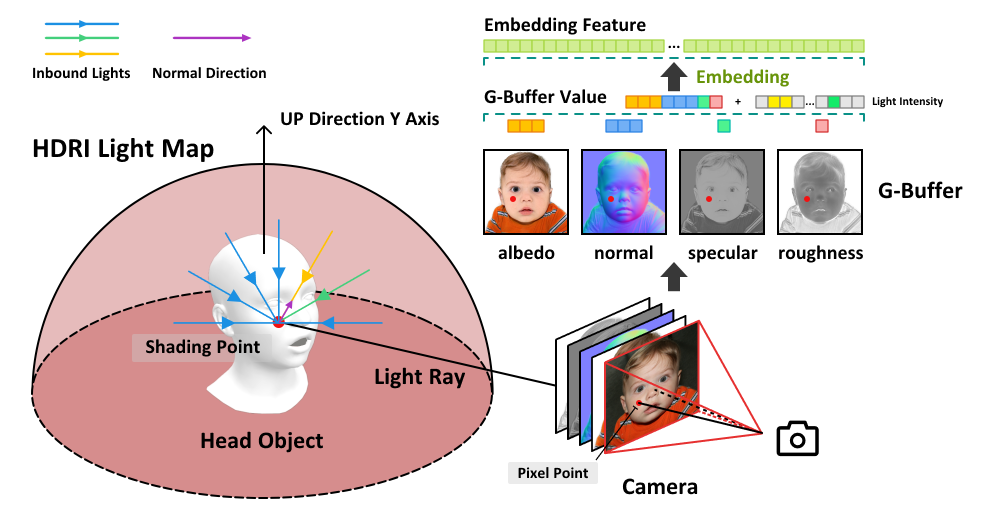}
    \caption{Shading process of each point. Each spatial point would be acquired all PBR texture values to form the G-Buffer and sampling inbound light from HRDI map to get the high dimensional feature as the input of neural network.}
    \label{fig:fig03}
\end{figure}

\subsection{Physics Based Neural Deferred Shading}
\label{sec:pbnds_method}
To approximate the rendering equation more precisely, our work redesigns previous neural deferred shaders~\cite{worchel_multi-view_2022} to incorporate PBR inputs (PBR materials, normal map, depth map, HDRI light map). 
% including PBR textures capturing physical material parameters, normal maps and HDRI light maps, as well as predicting an approximate shadow map. Fig.~\ref{fig:fig02} illustrates the overall training pipeline.
Due to the lack of paired samples for PBR inputs and real-world images, we first create a new dataset for training and evaluation; then train our neural deferred shader and shadow estimator by minimizing the difference between rendered and ground truth images. During inference, our model is capable of rendering an arbitrary G-buffer containing physically-based material and geometry information, under a given HDRI light-map, producing a photo-realistic image result.
\vspace{0.2em}

\noindent \textbf{Dataset creation.} We use the recent material and light predictor~\cite{kim_switchlight_2024} to create our dataset from images. This includes physically-based rendering (PBR) material textures, normal map, depth map, screen-space ambient occlusion (SSAO) map and image-based lighting (IBL) environment map, where the PBR textures are 2D texture maps consisting of albedo, specular, roughness, combined with normal map and depth maps to form the G-Buffer (see Sec. \ref{sec:NDR}). The light model is an equirectangular projection of a hemispherical HDR image that surrounds the object to provide the lighting. Moreover, calculating the outbound direction of each pixel requires the field-of-view (FOV) of the input image. We use a 3D Morphable Model (3DMM) to register input images and locate 3D landmarks on faces \cite{feng_learning_2021}, which combined with projected 2D landmarks and acquired depth map, allows to estimate the FOV. Using this process, we extend the human face dataset FFHQ \cite{karras_style-based_2019} into a new dataset called FFHQ256-PBR, consisting of 69,990 facial images with estimated PBR material textures, HDRI illumination and camera field-of-view. We manually remove failure cases (e.g.~where background removal affects the head itself). Our main experiments focus on human faces due to their importance in many applications; however other object types are also supported in our rendering pipeline. We therefore collect a second dataset based on a subset (1,000 scenes) of the synthetic BlenderVault dataset~\cite{litman2025materialfusion}. For this, we estimate the relevant parameters using the inverse rendering approach proposed in \cite{litman2025materialfusion}.
\vspace{0.2em}

% The material textures estimation removes image background which may cause partial loss of portraits, we remove these wrong samples manually.
% \begin{figure}
%     \centering
%     \includegraphics[width=0.8\textwidth]{Images/img03.png}
%     \caption{The FFHQ256-PBR human facial dataset consists of 69,990 real-world human facial images with estimated PBR material textures and IBL environment maps.}
%     \label{fig:fig03}
% \end{figure}

\noindent \textbf{Shading process.} We train the neural deferred shader to estimate the RGB color for each pixel, after which it can shade an arbitrary surface point during inference phase. Specifically, the shader processes each pixel of the G-Buffer separately (Fig.~\ref{fig:fig03}).
Texture and normal information is represented in screen space and taken from the corresponding pixel, while illumination information is sampled from rays on the hemisphere around the 3D point found by casting a ray from the pixel into the scene.
% through ray casting as shown in  where the imaging camera would cast a light ray to the shading point through the pixel point on the image plane. The pixel coordinates allow sampling texture information while uniform sampling light rays around the point on the hemisphere light map to provide the inbound light intensity $L_i$ and direction $\mathbf{l}$.

The neural shader's input is inspired by the classical rendering equation (eq.~\ref{eqn:eqn01}), and consists of three material features corresponding to $F$ term: albedo $\mathbf{a} \in [0, 1]^3$, specular $\mathbf{s}\in [0, 1]^1$ and roughness $\mathbf{r} \in [0, 1]^1$; three geometric terms (as unit vectors): normal $\mathbf{n} \in [-1, 1]^3$, inbound light direction $\mathbf{l}~\in~\mathbf{R}^3$ and outbound light direction $\mathbf{v} \in \mathbf{R}^3$; a single light feature: inbound light intensity $L_i \in (\mathbf{R}^+)^3$. 
The per-pixel values of $\mathbf{a}, \mathbf{n}, \mathbf{s}, \mathbf{r}$ form a geometry buffer (G-Buffer) that is concatenated with the $n$ sampled inbound light rays  \{$L_i(\mathbf{l}_1),L_i(\mathbf{l}_2),...,L_i(\mathbf{l}_n)$\} to form the input of a trainable neural shader with parameter $\theta$. Accordingly, the neural shading function is:
\begin{equation}
    \int_{\Omega} f_{\theta}(\mathbf{a}, \mathbf{n}, \mathbf{s}, \mathbf{r}, \mathbf{v}, L_{i}(\mathbf{l}) \langle \mathbf{n} \cdot \mathbf{l} \rangle) d\mathbf{l} \in [0, 1]^3, \mathbf{l} \in \Omega
    \label{eqn:eqn02}
\end{equation}
where $L_i(\mathbf{l}) \langle \mathbf{n} \cdot \mathbf{l} \rangle$ is the inbound light from upper hemisphere $\Omega^+$ similar to Eqn.~\ref{eqn:eqn01}. Thus the neural shader first regresses outbound light due to each inbound light direction for each pixel, then calculates the mean of these over the sampled inbound rays to get the final result. Thus neural shader learns to approximate the integral of the shading calculation from the reconstruction loss.
\vspace{0.3em}

% The architecture of the neural shader consists of two networks (see Fig~\ref{fig:fig03}): a diffuse network and a specular network. Both networks are multi-layer perceptrons (MLPs) consisting of 2 fully connected layers, using ReLU activation function. 
% As MLPs are ill-suited to learn functions with high spatial frequency \cite{jacot_neural_2018, tancik_fourier_2020}, we first embed the G-buffer and light intensity into a high-dimensional Fourier feature via positional embedding \cite{mildenhall_nerf_2020, worchel_multi-view_2022}. The diffuse network then processes the embedding features to produce the diffuse features. Note that since diffuse reflection is defined as isotropic, the view angle is not an input of this network. 

% The diffuse features are then concatenated with both the outbound direction $\mathbf{v}$ and the half direction $\mathbf{h}$ and fed to the specular network to gain view-dependent specular reflection, and produce an RGB color for the pixel. 

% \begin{figure}[htb]
%     \centering
%     \includegraphics[width=0.45\textwidth]{Images/img05.png}
%     \caption{Architecture of physics based neural deferred shader. The PBR input is transformed through the positional encoding \cite{tancik_fourier_2020} and fed to the diffuse network. The resulting feature vector is then concatenated with both the outbound direction $\mathbf{v}$ and half direction $\mathbf{h}$ and fed as input to specular network, producing a RGB color value.}
%     \label{fig:fig05}
% \end{figure}

\begin{figure}[t]
    \centering
    \includegraphics[width=0.65\textwidth]{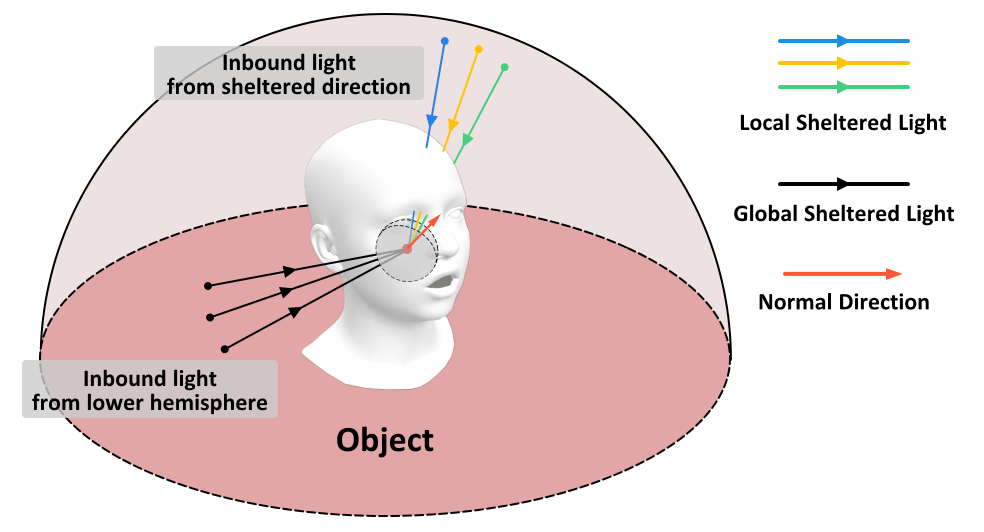}
    \caption{Localized shadowing model: The light rays denoted by black arrows come from the lower hemisphere and would be filtered by the cosine term in the rendering equation, whereas the light rays denoted by colorful arrows are occluded by local geometry resulting in localized shadowing.}
    \label{fig:fig04}
    \vspace{-0.5em}
\end{figure}

\noindent \textbf{Shadowing process.}
\label{sec:shadowing}
The architecture proposed above does not account for how inbound light rays may be affected by self-occlusions---for example, the top of the head may shelter light rays incident to a point below the eye, casting a shadow. Fig.~\ref{fig:fig04} illustrates this effect: light rays coming from the top side of the environment map are occluded by the top of the head. In practice, shadowing varies with differences in environment lighting: if the environment light map is even and smooth, screen space ambient occlusion (SSAO) can approximate the shadow effect well; conversely, if the environment light is highly anisotropic then the resulting hard shadows must be calculated by costly tracing of individual light paths. To calculate the shadow effect with a unified approach suitable for diverse light maps, we design a neural shadow estimator that learns to improve the shadowing result from our neural deferred shader. Specifically, a UNet-like neural network $f_\eta$ is designed to estimate the shadow map from the input of an unshadowed image $\hat{I}_{unshadow}$ and corresponding geometry information (normal map $I_{normal}$ and SSAO map $I_{ssao}$). The final shading result is the element-wise (Hadamard) product of the neural shader's output and the estimated shadow map.
The shadowing process is thus described by
\begin{equation}
\begin{gathered}
    \hat{I}_{shadow} = f_{\eta} (\hat{I}_{unshadow}, I_{normal}, I_{ssao}) \\
    \hat{I}_{rgb} = \hat{I}_{unshadow} \odot \hat{I}_{shadow}
\end{gathered}
\label{eqn:eqn03}
\end{equation}
where $\hat{I}_{\text{shadow}}$ is the pixel-wise shadow map, $\eta$ is parameters of the neural shadow estimator, $\hat{I}_{rgb}$ is the shadowed image.

\subsection{Loss Function}
\label{sec:loss}
%In our framework, solving the regression of pixel appearances formally equates to addressing the following minimization problem:
The proposed model is trained by solving the following minimisation for the neural network parameters $\theta$: 
\begin{equation}
    \mathop{\mathrm{argmin}}\limits_{\theta} L_{\text{appearance}}(\mathbf{a}, \mathbf{n}, \mathbf{s}, \mathbf{r}, \mathbf{v}, L_i)
    \label{eqn:eqn04}
\end{equation}
where $L_{\text{appearance}}$ compares the rendered pixel to the ground truth pixel at the corresponding position. $\mathbf{a}, \mathbf{n}, \mathbf{s}, \mathbf{r}, \mathbf{v}, L_i$ are albedo, normal, specular, roughness, outbound direction, inbound light respectively.
\vspace{0.2em}

More specifically, we train our model in two stages, with $L_{\text{appearance}}$ defined differently in each phase. In phase one, the model does not consider shadowing, thus $L_{\text{appearance}}^{(1)}$ is the shading loss defined by Eq.~\ref{eqn:eqn05}, which is an
% $L_{\text{appearance}}$ is composed of two terms, $L_{\text{shading}}$ and $L_{\text{shadowing}}$, that are optimized alternately in different epochs, ensuring their performance are improved synchronously. 
$L_1$ loss calculating the average distance between prediction and ground truth for all foreground pixels, $N_S$ is the number of sampled pixels, and $C$ is the RGB value of pixel.
\begin{gather}
    L_{\text{appearance}}^{(1)} = \frac{1}{N_S}\sum_{i=1}^{N_S}\| C_i - \hat{C}_i \|_1 \label{eqn:eqn05} \\
    L_{\text{appearance}}^{(2)} = \| I_{rgb} - \hat{I}_{rgb} \|_1 \label{eqn:eqn06}
\end{gather}

In phase two, the neural deferred shader is frozen to produce an unshadowed image, which is fed to the shadow estimator through concatenating with the normal map (see Sec.~\ref{sec:shadowing}), yielding the shadowed image~$\hat{I}_{rgb}$. An L1 loss $L_{\text{appearance}}^{(2)}$ (Eq.~\ref{eqn:eqn06}) calculates the residual between the shadowed image $\hat{I}_{rgb}$ and ground truth image $I_{rgb}$ for all pixels.

\subsection{Implementation Details}
\label{sec:implementation}
% We train our model in two stages: the shading stage and the shadowing stage. In the shading stage, the model learns to regress the color for each point irrespective of local occlusion; in the shadowing stage the model learns to regress the shadow map; the final rendering result is the pixel-wise product of their output. % this was already stated above
We train each stage for 40 epochs, which balances efficiency and performance. We use the Adam optimizer \cite{kingma_adam_2015}, with parameters $\beta_1=0.9$, $\beta_2=0.999$, and a learning rate of $5\times 10^{-5}$. We randomly sample 8,192 foreground pixels of each image to form each training batch; this provides enough diversity to avoid over-fitting as the neural deferred shading works on a per-pixel basis and is agnostic to the overall scene structure. We uniformly sample 128 inbound light rays from the HDRI light map as input to the shader in each training step.
Training converges within one day on a single NVIDIA GeForce RTX 3090 GPU. During inference, the physically based neural deferred shader can render an image in 1 second per frame.

\section{Experiments}
\label{sec:experiment}
We conducted comprehensive experiments to evaluate the performance of our neural deferred shading approach, comparing qualitatively and quantitatively against three baselines. The baseline models include classical shading models (the empirical Blinn-Phong model and the physics-based GGX model) as well as a recent learning-based model, neural deferred shader~\cite{worchel_multi-view_2022}.
% We compare performance qualitatively and quantitatively between our model and the baselines.
\begin{figure}[htb]
\centering
% Left figure
\begin{minipage}[b]{0.48\linewidth}
    \centering
    \includegraphics[width=\linewidth]{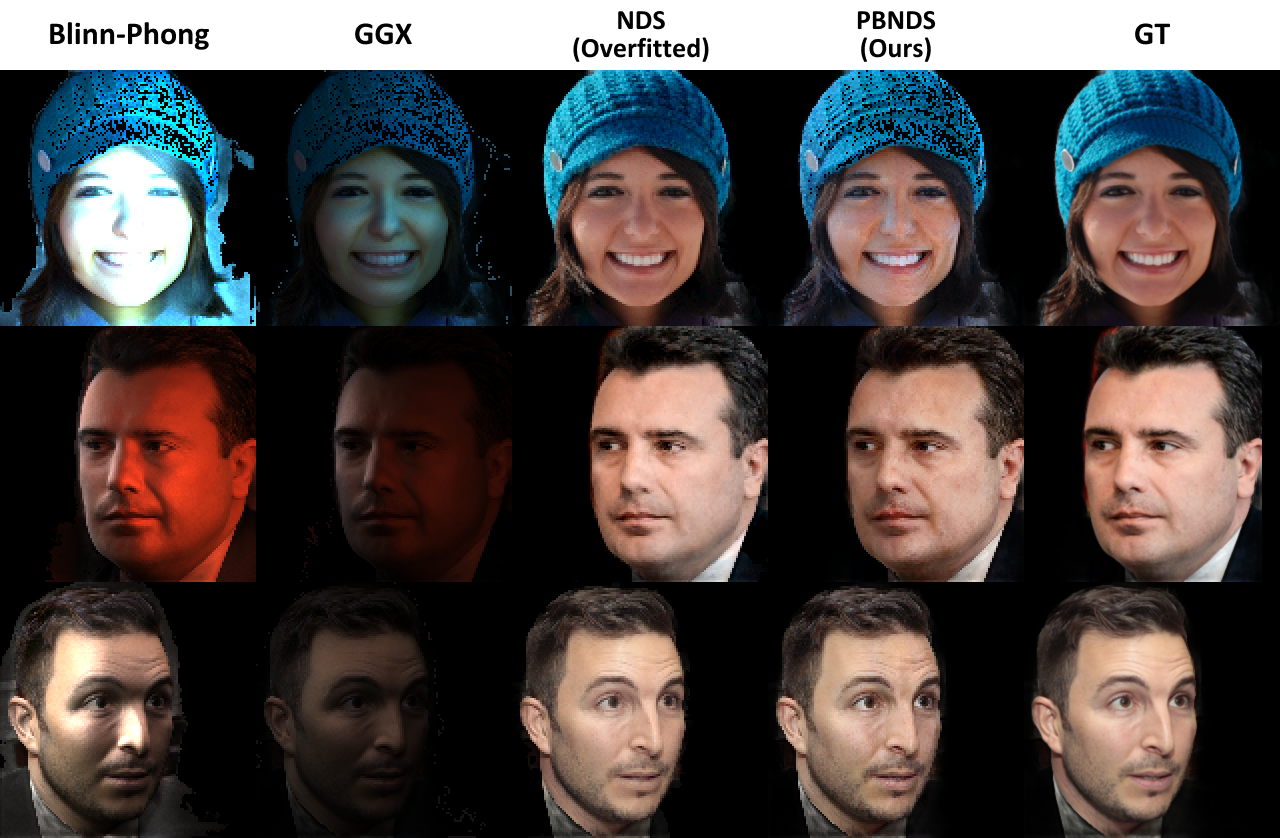}
\end{minipage}
% Right figure
\begin{minipage}[b]{0.48\linewidth}
    \centering
    \includegraphics[width=\linewidth]{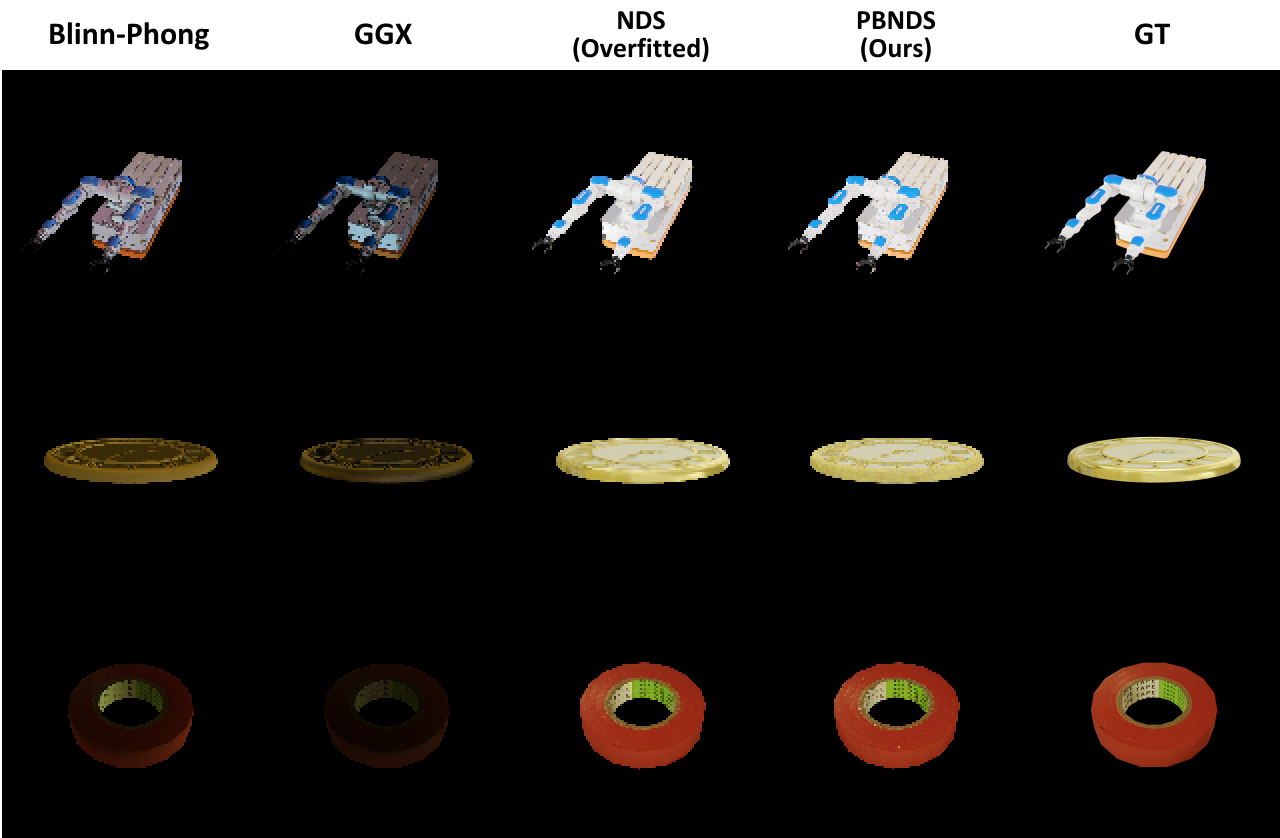}
\end{minipage}
\caption{Quality comparison between different shading models. \textbf{Blinn-Phong}: Blinn-Phong shading model; \textbf{GGX}: Trowbridge-Reitz GGX model; \textbf{NDS}: Neural deferred shader; \textbf{PBNDS}: Physics based neural deferred shader; \textbf{GT}: Ground truth image. Our model can reconstruct the ground truth scene with generalization and keep the photo-realism maximally through capturing the correct light-surface interaction.}
\label{fig:fig05}
\end{figure}
\vspace{-1.5em}

\subsection{Qualitative Evaluation}
\label{sec:qualit_eval}
We visually compare shading results produced by different shading models from similar inputs. We evaluate examples from two different datasets, a separate test set drawn from the FFHQ256-PBR for facial data and a subset of BlenderVault~\cite{litman2025materialfusion} for other object types. Figure \ref{fig:fig05} illustrates the results for both datasets, comparing our neural shader with the classical Blinn-Phong~\cite{blinn_models_1977} and GGX~\cite{walter_microfacet_2007} models. Our physical based neural deferred shader (PBNDS) reconstructs the image photo-realistically from the estimated materials and illumination, and the overall hue and light reflection are more realistic than classical models. In contrast, the Blinn-Phong model is an empirical model where the light reflection cannot produce a photorealistic result.
The GGX model is based on micro-facet theory which requires the additional alignment for the incident light intensity to actualize the energy conservation, and is not suited for the estimated illumination application. We also compare our model to the state-of-art neural deferred shading (NDS) \cite{worchel_multi-view_2022}, and show that our approach can achieve similar quality to it. This is notable since ours is a generalized (scene-agnostic) model  where NDS overfits a neural network separately to each scene, since it takes the scene-dependent point coordinates as an input.
\begin{figure}[htb]
\centering
% Left figure
\begin{minipage}[b]{0.49\linewidth}
    \centering
    \includegraphics[width=\linewidth]{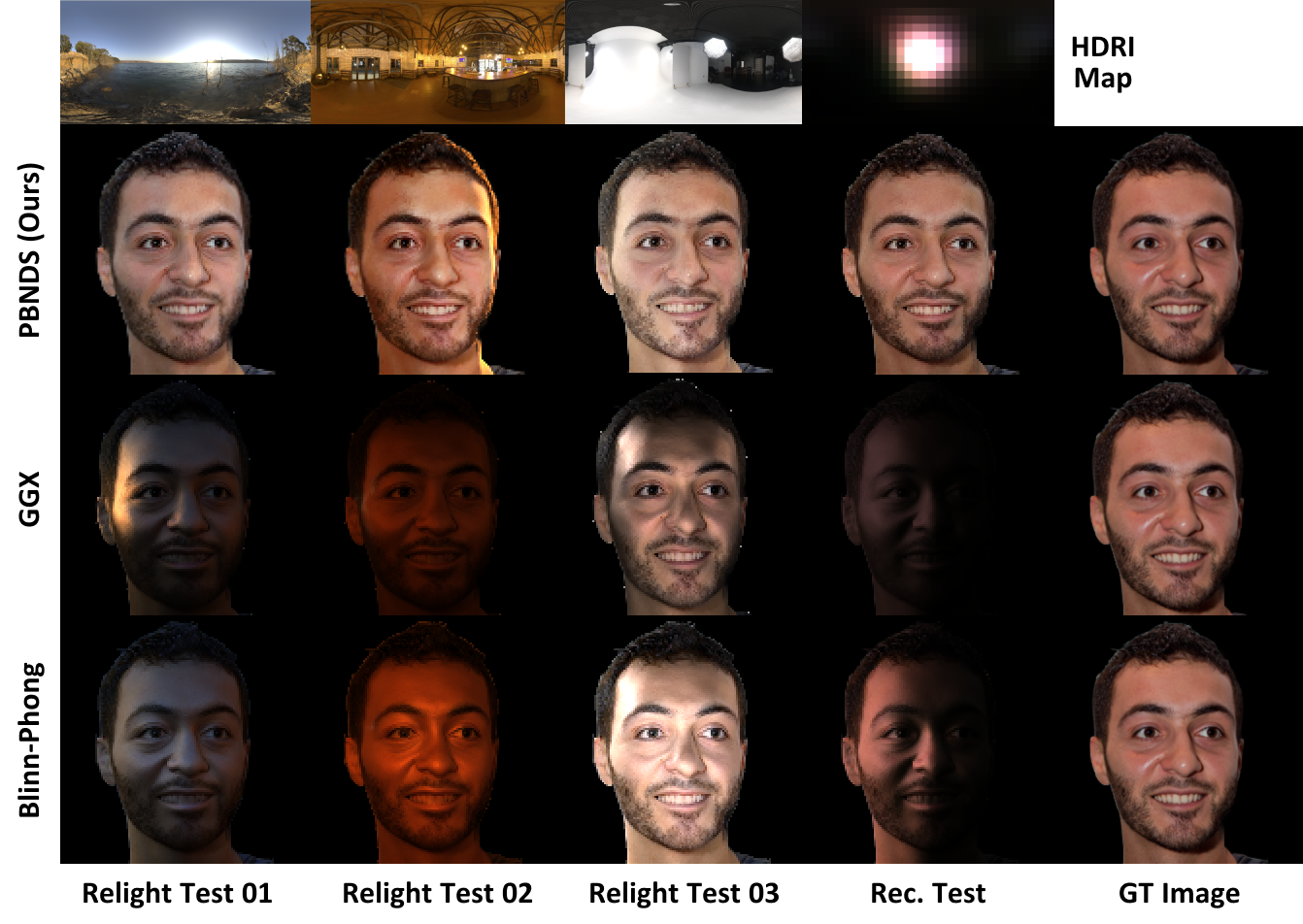}
    \label{fig:left}
\end{minipage}
% Right figure
\begin{minipage}[b]{0.49\linewidth}
    \centering
    \includegraphics[width=\linewidth]{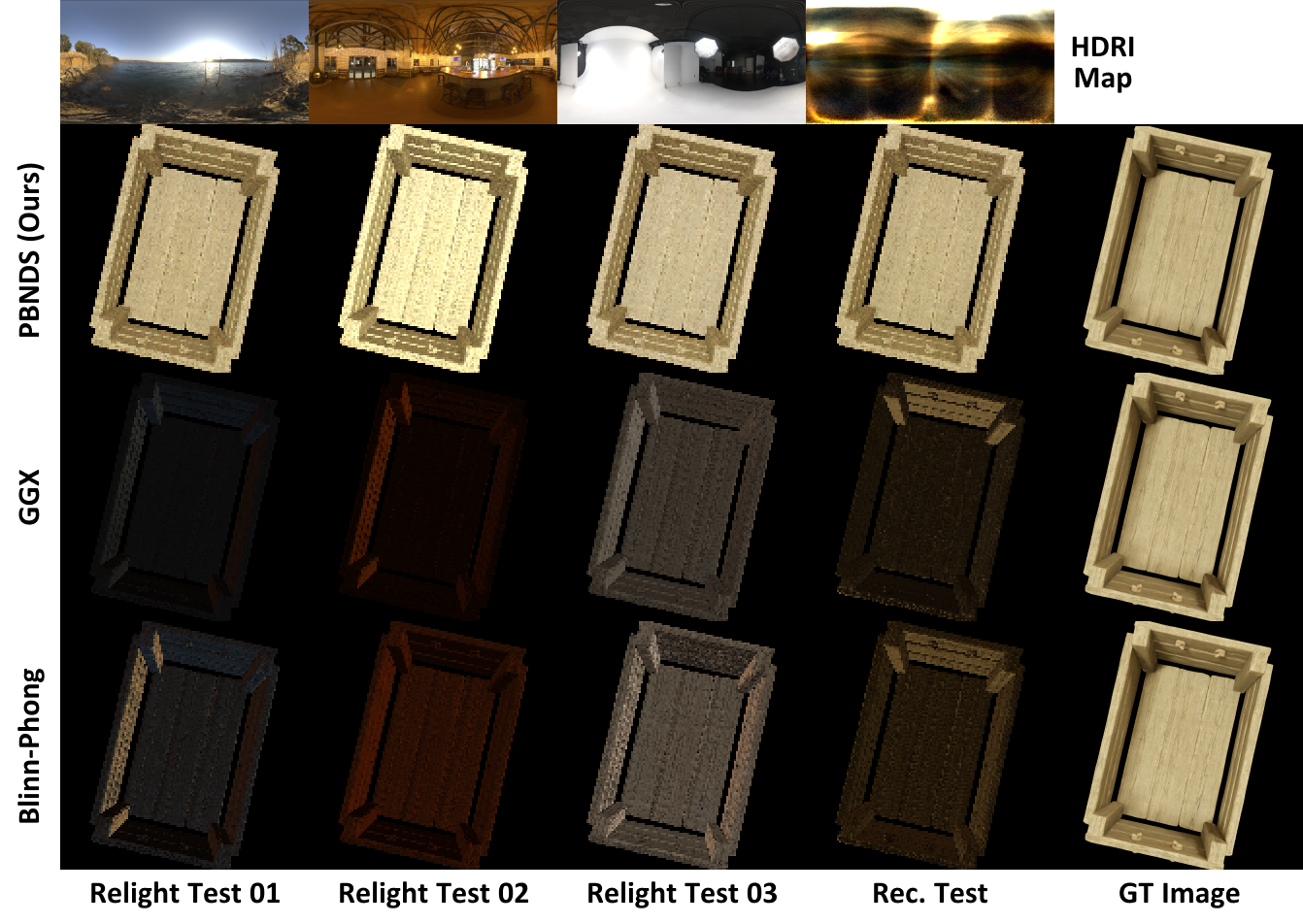}
    \label{fig:right}
\end{minipage}
\caption{Rendering results with different HDRI maps. The first row shows HDRI maps used to relight the original head; the other rows show the resulting rendered images after relighting from different shading methods. 
The results show our neural shader allows the environment map to realistically influence the shading of the head.}
\label{fig:fig07}
\end{figure}

We next show how our model enables relighting images, by using recorded HDRI maps to render the image with different illumination. Specifically, we use our model trained from the previous section for both types of data but replace the HDRI map during rendering. All the new HDRI maps are captured in real-world scenes. We compare the relighting rendering results to classical rendering methods (Blinn-Phong, GGX) in Fig.~\ref{fig:fig07}. Our model correctly recovers the light-surface interaction within the scene with high fidelity, whereas the classical models do not work well in this inverse setting. Compared to rendering the scene with estimated illumination, the relighting test requires our model to adapt to real-world illumination inputs, which may lie outside the distribution of estimated illumination seen during training.
% , thanks to the positional encoding \cite{mildenhall_nerf_2020} which encodes the input to the frequency domain and let the neural network learn the fundamental Fourier feature, gaining the generalization and decomposition for our model.
\vspace{-0.5em}

\subsection{Quantitative Evaluation}
\label{sec:quant_eval}
We also compare the rendered images to the ground-truth quantitatively. 
We conduct 500 tests in each experiment to measure the performance of each shading model, using data from the test set of FFHQ256-PBR and BlenderVault~\cite{litman2025materialfusion}. To measure the realism of the shading result, we use Learned Perceptual Image Patch Similarity (LPIPS) \cite{zhang_unreasonable_2018} and Fréchet inception distance (FID) \cite{heusel_gans_2017} as metrics; lower LPIPS and FID scores indicate better realism comparing ground truth and real world data. We also measure the Peak signal-to-noise ratio (PSNR) and Structural similarity index measure (SSIM), where higher PSNR or SSIM score indicates better reconstruction performance. As there is no paired ground-truth data for the relighting experiment, we only use FID score for evaluation.

% \begin{table}[h]
%   \centering
%   \caption{Quantitative evaluation of shading models}
%   \resizebox{0.6\linewidth}{!}{
%   \begin{tabular}{l l l l l}
%     \toprule
%     Models          & MSE $\downarrow$ & PSNR $\uparrow$ & SSIM $\uparrow$ & LPIPS $\downarrow$ \\
%     \midrule
%     Blinn-Phong     & 0.0775           & 11.90           & 0.6227          & 0.1809 \\
%     GGX             & 0.1267           & 9.11            & 0.4894          & 0.2726 \\
%     NDS(Overfitted) & \textbf{0.0050}  & \textbf{29.68}  & 0.8847          & 0.0617 \\
%     Ours            & 0.0066           & 24.61           & \textbf{0.9191} & \textbf{0.0318} \\
%     \bottomrule
%   \end{tabular}
%   }
%   \label{tab:tab01}
% \end{table}

\begin{table}[tb]
  \centering
  \caption{Quantitative evaluation for shading/relighting experiments}
  \resizebox{\textwidth}{!}{
  \begin{tabular}{@{}clcccccccccc@{}}
    \toprule
      & & \multicolumn{2}{c}{\small MSE $\downarrow$} & \multicolumn{2}{c}{\small PSNR $\uparrow$} & \multicolumn{2}{c}{\small SSIM $\uparrow$} & \multicolumn{2}{c}{\small LPIPS $\downarrow$} & \multicolumn{2}{c}{\small FID $\downarrow$} \\
      \cmidrule(r){3-4} \cmidrule(r){5-6} \cmidrule(r){7-8} \cmidrule(r){9-10} \cmidrule{11-12} 
      & & \small FP & \small BV & \small FP & \small BV & \small FP & \small BV & \small FP & \small BV & \small FP & \small BV \\
    \midrule
      \multirow{4}{*}{\centering \textbf{SD}}
        & Blinn-Phong & 0.078 & 0.044 & 11.90 & 14.94 & 0.623 & 0.784 & 0.181 & 0.080 & 0.244 & 0.239 \\
        & GGX         & 0.127 & 0.055 & 9.11  & 13.35 & 0.489 & 0.753 & 0.273 & 0.114 & 0.368 & 0.227 \\
        & NDS (Overfitted) & \textbf{0.005} & \textbf{0.001} & \textbf{29.68} & \textbf{38.53} & 0.885 & \textbf{0.988} & 0.062 & 0.027 & 0.179 & \textbf{0.037} \\
        & PBNDS (Ours) & 0.007 & 0.002 & 24.61 & 29.47 & \textbf{0.919} & 0.936 & \textbf{0.032} & \textbf{0.025} & \textbf{0.056} & 0.113 \\
    \hline
    \rule{0pt}{8pt}
      \multirow{3}{*}{\textbf{RE}}
        & Blinn-Phong & - & - & - & - & - & - & - & - & 0.331 & 0.163 \\
        & GGX         & - & - & - & - & - & - & - & - & 0.556 & 0.162 \\
        & PBNDS (Ours) & - & - & - & - & - & - & - & - & \textbf{0.090} & \textbf{0.117} \\
    \bottomrule
  \end{tabular}}
  \label{tab:tab01}
  \parbox{0.9\linewidth}{\tiny \textbf{SD:} Shading experiment; \textbf{RE:} Relighting experiment; \textbf{FP:} FFHQPBR dataset; \textbf{BV:} BlenderVault dataset}
\end{table}

As shown in Tab.~\ref{tab:tab01}, our model exceeds classical shading models significantly in all metrics for both shading and relighting experiments, as the training process can adapt our model for the input with different data domain, especially for rendering the real world scene whose materials and illumination are estimated and not calibrated for classical models. Our model also outperforms the overfitted neural deferred shader model \cite{worchel_multi-view_2022} in SSIM, LPIPS and FID for FFHQPBR dataset, and achieves competitive results in the less perceptually-relevant MSE and PSNR metrics for all datasets. We suggest this is caused by pixel-wise overfitting causing high-frequency noise especially for real-world images, that can be alleviated by the global shadowing process.
\begin{table}[tb]
  \centering
  \begin{minipage}[t]{0.58\linewidth}
  \centering
  \caption{Comparison of training with different model components}
  \label{tab:tab02}
  \resizebox{\linewidth}{!}{
  \begin{tabular}{lccccc}
    \toprule
        & MSE $\downarrow$ & PSNR $\uparrow$ & SSIM $\uparrow$ & LPIPS $\downarrow$ & FID $\downarrow$ \\
    \midrule
    Default Config.  & 0.0046           & 24.4107        & 0.9167          & 0.0337           & 0.0607\\
    + Perc. Loss     & 0.0052           & 24.0763        & \textbf{0.9298} & 0.0273           & 0.0441\\
    + SDO (w/o AO)   & 0.0041           & 24.6948        & 0.9194          & 0.0319           & 0.0559\\
    + SDO (AO)       & \textbf{0.0039}  & \textbf{25.0053} & 0.9247        & \textbf{0.0262}  & \textbf{0.0240} \\
    \bottomrule
  \end{tabular}}
  \hspace{0.1\linewidth}
  \end{minipage}
  \begin{minipage}[t]{0.39\linewidth}
  \centering
  \caption{Comparison of different training batch size}
  \label{tab:tab03}
  \resizebox{\linewidth}{!}{
  \begin{tabular}{cccc}
    \toprule
        & PSNR $\uparrow$ & GPU Mem. & FLOPs(G) \\
    \midrule
    2048        & 21.74 & 2.98 & 0.64 \\
    4096        & 22.81 & 5.96 & 1.28 \\
    8192        & 24.83 & 11.91 & 2.56 \\
    Full Res.   & 25.81 & 23.82 & 5.12 \\
    \bottomrule
  \end{tabular}}
  \end{minipage}
\end{table}

\subsection{Ablation study}
We conducted an ablation study to evaluate the importance of the different components of our neural shading pipeline, following the same approach as in Sec.~\ref{sec:pbnds_method}. We test our shading model with different configurations, each test follows a design similar to Sec.~\ref{sec:quant_eval}. The default configuration of our neural deferred shader lacks a shadow estimator, and uses a single reconstruction loss (as in~Sec.\ref{sec:loss}), ReLU activation function, positional encoding \cite{mildenhall_nerf_2020}; other tests add specific components to this default setting. We add a perceptual loss \cite{leibe_perceptual_2016} in the second test and our shadowing process in the third.
We see in Tab.~\ref{tab:tab02} that the perceptual loss improves results of perceptual metrics (SSIM, LPIPS, FID) but downgrades the reconstructive metrics (MSE, PSNR); the shadowing process without ambient occlusion (AO) can balance on both aspects, with AO the performance can be further improved to achieve the best result on most metrics with only slight degradation on SSIM metric, thus we use the default setting with shadowing process (AO) as the final design. We also compare the training result of different batch size and find 8,192 is the best choice to balance the model performance and training efficiency (see Tab.~\ref{tab:tab03}).

% We tried to enhance the capturing ability of high frequency features through using SIREN activation function \cite{} and Gaussian Fourier Feature Transformation (GFFT) encoding method \cite{tancik_fourier_2020} in test 4 and test 5, which are proven the useless for all metrics.

% Fig~\ref{fig:fig08} illustrates the qualitative comparison of all tests in the ablation study. The default setting reconstructs the image in pixel level without global supervision, resulting in the hard boundary in the middle of the face; Perceptual loss works globally, reducing the hard boundary but losing some shadowing effect; The global shadowing process balances the overall effect, producing the best outcome.

%The SIREN activation leads to more noisy result and GFFT encoding decays the shading completely.

% \begin{figure}[htb]
%     \centering
%     \includegraphics[width=0.6\textwidth]{Images/img09.png}
%     \caption{Comparison of shading results with different experiment configurations in the ablation study. The default setting renders image pixel by pixel resulting in some wrong result; Adding perceptual loss fixes it but lacks the shadow; The shadowing process restores them and gets the balanced result.}
%     \label{fig:fig08}
% \end{figure}

\section{Conclusion}
\label{sec:conclusion}
We have presented a new method for photo-realistic rendering by using physics-based neural deferred shading, demonstrating improved performance compared to existing methods. We showed our method can be used for effective relighting and rendering. We also proposed a novel human facial dataset with PBR material textures, FFHQ256-PBR, which can be used for future research on real-world rendering.

Our method does not explicitly model the domain shift between synthetic, estimated and real-world data; although the relighting experiment proves it can handle real-world HDRI input, manually created 3D models still might be unsuitable for our shading model due to the, which could be addressed in future work by incorporating a domain adaptation process. 

% The shadow estimator occasionally leads to overly dark renderings; this could be improved by using real-time ray tracing in future work.

\bibliographystyle{splncs04}
\bibliography{PBNDS}
\end{document}